\documentclass{article}
\usepackage{spconf}
\usepackage{amsmath}
\usepackage{amssymb}
\usepackage{graphicx}
\usepackage{booktabs}
\usepackage{enumitem}
\usepackage{dashrule}
\setlist{nosep, leftmargin=14pt}
\usepackage{url}

\title{ContactCut: A Partition-Aware Contact Selection Framework for False-Merge Correction in Connectomics}

\name{Byunghyun Ko$^{1}$, Won-Ki Jeong$^{2}$, Chaoli Wang$^{1}$%
\thanks{This work has been submitted to the IEEE for possible publication.
Copyright may be transferred without notice, after which this version may no longer be accessible.}}
\address{
$^{1}$Department of Computer Science and Engineering, University of Notre Dame, Notre Dame, IN, USA\\
$^{2}$Department of Computer Science and Engineering, Korea University, Seoul, Korea
}

\begin{document}
\maketitle

\begin{abstract}
Localized false-merge correction in connectomics requires converting an identified error into an executable native 3D partition. Still, neither intermediate correction predictions nor seeded graph solvers alone determine which native source--sink constraints should define that partition. We formulate this problem as a 3D-native partition-aware contact selection problem: selecting candidate contacts based on the complete partitions they induce. Our solution comprises two components: ContactCut (CC)-Auto, which uses learned single-contact selection, and ContactCut (CC)-Search, which enables bounded paired-contact search, generates compact alternatives, and applies learned panel ranking. On historical two-way false-merge corrections from MICrONS and FlyWire, CC-Auto improves failure-aware 3D partition (FA-3D Part.) mIoU by 5.9 and 3.4 points over the strongest graph-based controls and by 12.5 and 10.0 points over the best learned baselines. A MICrONS-trained ranker transfers to FlyWire with negligible degradation. CC-Search expands oracle reachability and prioritizes near-oracle alternatives. The code is available at \url{https://github.com/bentake/ContactCut}.
\end{abstract}

\section{Introduction}
\label{sec:intro}
Advances in large-scale electron microscopy (EM) connectomics enable reconstruction of neural circuits from cortical volumes to complete adult brains~\cite{microns2025,dorkenwald2022flywire}, yet residual segmentation errors still require substantial proofreading. \emph{False merges} are particularly damaging because a single erroneous fusion can corrupt the morphology, attributed connectivity, and downstream interpretation of multiple neurons~\cite{matejek2019}. They are also harder to correct than \emph{false splits}: reconnecting separated fragments can often be posed as a linkage decision, whereas correcting a false merge requires recovering a 3D partition of an already fused object. Although human-level false-merge identification has been reported, comparable correction performance has not been demonstrated~\cite{brown2026}. Systems like FlyWire, CAVE, and NeuTu initiate splitting from user-provided points or seeds associated with the erroneous region~\cite{dorkenwald2022flywire,cave2025,zhao2018}. We therefore study the subsequent correction step, given a localized false-merge cue.

Prior work has addressed false-merge correction through volumetric prediction, skeleton-based localization, candidate proofreading, and morphological or graph-based reasoning~\cite{zung2017,dmitriev2018,haehn2018,li2020,troidl2025,chen2026morphology}. Zung et al.~\cite{zung2017} predicted correction masks, Dmitriev et al.~\cite{dmitriev2018} used skeletal structure to localize and guide corrections, Troidl et al.~\cite{troidl2025} recovered neuron labels from learned point-cloud affinities, and Brown et al.~\cite{brown2026} predicted merge-correction masks followed by hull reconstruction, seed sampling, and graph splitting. Guided proofreading~\cite{haehn2018} generates seeded candidate splits, placing random watershed seed pairs on opposite sides of a 2D segment, but scores each only by its local boundary using an image-patch classifier. Reinforced proofreading learns policies for corrective actions~\cite{nguyen2021,schmidt2024}, and other systems use neuronal morphology or structural rules to identify and remove segmentation errors~\cite{li2020,celii2025,huang2025}. These methods provide evidence for how a false merge should be separated, but their intermediate predictions do not necessarily determine the native partition. Brown et al., for example, reported that high-IoU correction masks can still yield poor supervoxel splits after reconstruction and splitting.

On the other hand, interactive systems such as FlyWire, CAVE, NeuTu, and Neu3 produce executable edits once suitable labels, seeds, or split locations are supplied~\cite{dorkenwald2022flywire,cave2025,zhao2018,hubbard2020,xenes2022neuvue}. CAVE expands terminal sets to reduce sensitivity to precise label placement~\cite{cave2025}, and natural-image proposal methods such as CPMC~\cite{carreira2012cpmc} automatically place seeds and rank the resulting constrained min-cut segments, though for class-agnostic recall rather than a single executable partition of a merged object. Near a localized cue, however, several native contacts could serve as source--sink constraints, and different choices can induce different whole-object partitions even when the graph, edge weights, and downstream solver are fixed. We therefore formulate localized false-merge correction as \emph{partition-aware contact selection}: selecting native contact constraints based on the partitions they induce. 

We propose \textbf{ContactCut}, a novel 3D-native partition-aware contact-selection framework for localized false-merge correction that operates directly on the supervoxel graph. Given a falsely merged object and a localized interface cue, ContactCut enumerates nearby native contacts, assigns each contact's endpoint supervoxels to the corresponding terminals, and evaluates the resulting complete 3D partition using a fixed-seeded solver without relying on image appearance. \textbf{ContactCut (CC)-Auto} learns to rank single-contact candidates from ground-truth (GT)-free partition structure, while \textbf{ContactCut (CC)-Search} expands the candidate space through bounded paired-contact search, compact alternative generation, and learned panel ranking. Our contributions are:
\begin{itemize}
    \item We formulate localized false-merge correction as 3D-native partition-aware contact selection. Controlled experiments isolate contact selection from the downstream solver and show that selected terminals remain beneficial when reused with graph cut, random walker, and watershed on both evaluated datasets.
    \item We introduce a lightweight partition-aware ranker for automatic single-contact selection, along with a bounded paired-contact search and a learned ranker for compact alternative panels. Both learned rankers transfer from MICrONS to FlyWire without retraining.
    \item We evaluate on genuine historical false-merge corrections from MICrONS and FlyWire under a common native two-way protocol, where CC-Auto achieves the strongest automatic correction performance against other methods.
\end{itemize}

\vspace{-0.25cm}
\section{Method}
\vspace{-0.15cm}
\label{sec:method}

\begin{figure}[t]
    \centering
    \includegraphics[width=0.9\columnwidth]{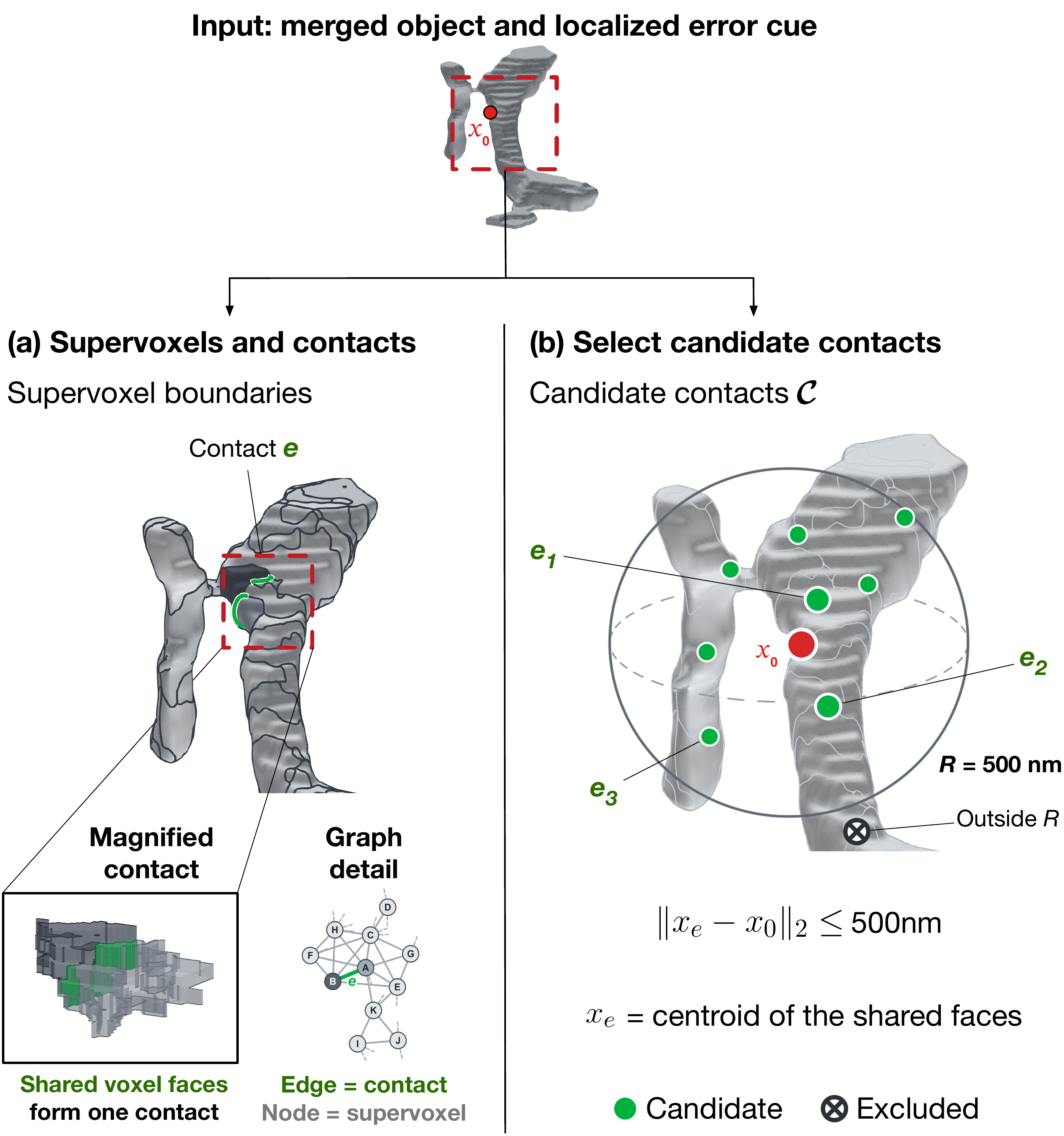}
    \vspace{-0.5cm}
    \caption{Supervoxel contacts and candidate selection.
(a) Supervoxels correspond to graph nodes. The voxel faces shared by two supervoxels form a contact, represented by a graph edge. (b) Candidate contacts have centroids within 500\,nm of the localized error cue $x_0$ (red). Shown candidate centroids are green. The crossed marker denotes an excluded contact.
}
    \vspace{-0.4cm}
    \label{fig:candidateselection}
\end{figure}

Let $G=(V, E)$ be the native supervoxel graph, with supervoxels as nodes and contacts $e=(u,v)\in E$ between supervoxels sharing voxel faces. Contact capacity $c_e$ is the shared-face count, and centroid $x_e$ is the arithmetic mean of all shared-face centers in physical coordinates. Given a localized error cue $x_0$, we enumerate $\mathcal C=\{e\in E:\|x_e-x_0\|_2\le500\,\mathrm{nm}\}$. We use 500\,nm, the smallest tested radius retaining full candidate coverage in validation. For each candidate contact, we assign supervoxels $u$ and $v$ to opposing source and sink terminals. A seeded $s$--$t$ minimum cut on the full graph then produces the whole-object partition $P_e$. Terminal capacities exceed the total native-edge capacity, enforcing these assignments. Figure~\ref{fig:candidateselection} shows the supervoxel contact representation and candidate selection. 

\noindent\textbf{CC-Auto.}\ 
Let $\mathcal T\subseteq\mathcal C$ contain candidates yielding two connected output sides, if any exist. Otherwise, retain those with the fewest extra connected components. For each induced partition $P_e$, let $f_s(P_e)$ be the native-voxel fraction assigned to its smaller side. Let $f_c(P_e)$ be the cut contact size: the sum of $c_{uv}$ over all graph edges whose endpoints lie on opposite output sides. The smaller-side fraction discourages the isolation of negligible fragments, while the cut contact size favors separation through narrow connections. These complementary structural preferences do not require equal-sized output neurons. Within $\mathcal T$, CC-Auto ranks candidates by
\begin{equation}
\vspace{-0.15cm}                                                                       
\begin{aligned}
\phi(P_e)
&=
\begin{bmatrix}
z(f_s(P_e))\\[0.6mm]
-z(\log(1+f_c(P_e)))
\end{bmatrix},\\[1mm]
e^\star
&=\arg\max_{e\in\mathcal T}\theta^\top\phi(P_e).
\end{aligned}
\vspace{-0.1cm}
\label{eq:auto}
\end{equation}
where $z(\cdot)$ subtracts the within-tier median and divides by the median absolute deviation (MAD), with values clipped to $\pm8$ and set to zero when the MAD vanishes. For each historical correction in the training split (TRAIN), we generate candidate partitions from the pre-edit graph and score them against the recorded two-descendant reference using 3D Part.\ mIoU. Within each operation's tier, candidate pairs differing by at least 0.02 mIoU provide winner--loser feature differences $\phi(P_{\rm win})-\phi(P_{\rm lose})$, with the higher-mIoU partition designated the winner. We retain at most 4096 pairs per operation and give each contributing operation equal total weight. An $L_2$-regularized pairwise logistic fit learns the two coefficients $\theta$. Component-grouped fivefold TRAIN cross-validation selects $C=1$. The coefficients calibrate the trade-off between structural preferences. Inference uses only the pre-edit graph and cue, without raw EM appearance or GT.

\noindent\textbf{CC-Search.}\ CC-Search expands the reachable correction space by using two contacts jointly to define source--sink constraints. We use a pair budget $B=45$, chosen in the saturation region of a validation sweep. If more than $B$ unordered contact pairs are available, we sample $B$ without replacement using a fixed random seed. Otherwise, we evaluate all pairs. We evaluate both relative source--sink orientations, skip conflicting assignments, and canonicalize/deduplicate the resulting partitions up to side swap. The single-contact base $P_{\rm H}$ is the deterministic, non-learned H\_FIXED anchor, which selects the candidate maximizing $s_{\rm H}(P_e)=10^9f_{\rm bc}(P_e)-10^5f_{\rm nextra}(P_e)+10^2f_s(P_e)-10^{-3}f_c(P_e)$, where $f_{\rm bc}$ indicates that both sides are connected and $f_{\rm nextra}$ counts extra connected components. On all evaluated operations, the first two terms select the same connectivity tier $\mathcal T$ as CC-Auto. Within $\mathcal T$, H\_FIXED uses the fixed trade-off $10^2f_s-10^{-3}f_c$. From the paired alternatives, greedy facility location~\cite{nemhauser1978} selects $K=5$ diverse alternatives, a compact inspection budget chosen at a diminishing-returns point in a validation sweep. Let $\mathcal A$ denote the deduplicated paired alternatives. Using similarity $s(P,Q)=1-d(P,Q)$, it greedily maximizes
$F(S)=\sum_{Q\in\mathcal A}\max_{P\in S\cup\{P_{\rm H}\}}s(P,Q)$
subject to $|S|\le K$, where the label-invariant voxel disagreement is
\begin{equation}
\vspace{-0.15cm}
\begin{aligned}
d(P,Q)
&=\frac{\min\{\delta(P,Q),\,M-\delta(P,Q)\}}{M},\\[1mm]
\delta(P,Q)
&=\sum_{\ell\in V}m_\ell\mathbf 1[P(\ell)\ne Q(\ell)].
\end{aligned}
\label{eq:search_distance}
\vspace{-0.1cm}
\end{equation}
Here $m_\ell$ is the native voxel count of supervoxel $\ell$ and $M=\sum_\ell m_\ell$. PairLogit, a pairwise logistic ranker~\cite{bradley1952}, orders the facility-selected alternatives. Its 58-dimensional feature vector $\psi(P)$ concatenates two versions of 29 GT-free descriptors of partition structure, interface distance, and dissimilarity to $P_{\rm H}$: one standardized using TRAIN means and standard deviations, and one normalized by median/MAD across the panel's alternatives and clipped to $\pm5$. The ranking score is
\begin{equation}
\vspace{-0.08cm}
r(P)=w^\top\psi(P).
\label{eq:pairlogit}
\vspace{-0.04cm}
\end{equation}
We train on 79 MICrONS TRAIN panels. An alternative is labeled positive if its reference mIoU is within 0.05 of the best value in the H\_FIXED-plus-B45 pool, and negative otherwise. Only positive--negative pairs within each panel contribute feature differences to the $L_2$-regularized logistic fit ($C=1$), yielding 172 training pairs with equal total weight per contributing panel. The frozen model reorders alternatives on MICrONS TEST and FlyWire without changing panel membership, and the base remains available.

Best-single is the reference-best single-contact partition. B45-pair is the reference-best partition in the H\_FIXED-plus-B45 pool, and Best-in-$K5$ is the reference-best member of the base-plus-five panel. NEAR05@$k$ is the fraction of operations for which the base or first $k$ ranked alternatives reach within 0.05 mIoU of the B45-pair optimum.

\vspace{-0.25cm}
\section{Experiments and Results}
\vspace{-0.15cm}
\label{sec:experiment_and_results}

\noindent\textbf{Implementation details.}\
We evaluated historical two-way false-merge corrections from MICrONS minnie65~\cite{microns2025} and FlyWire FAFB~\cite{dorkenwald2022flywire}. Pre-edit objects and their two post-edit descendants define the historical reference partitions, so our evaluation measures agreement with recorded corrections rather than globally verified neuron identity. The proofreading interface point provides the localized cue $x_0$. MICrONS uses dependency-component-disjoint TRAIN/VAL/TEST splits (328/142/754), while FlyWire follows the official CB2 TRAIN/TEST split (20,402/2,969; no VAL). ContactCut hyperparameters were fixed before TEST: $R=500$\,nm, $B=45$, and $K=5$ were selected on MICrONS VAL, CC-Auto regularization was selected by component-grouped TRAIN cross-validation, and other settings were fixed. All methods were evaluated as native two-way partitions using the same frozen scorer. Random and Geometric methods use random or PCA-derived planes through the interface. NC-RW and NC-WS denote nearest-contact random walker and seeded watershed/maximum-spanning forest, respectively. PAT~\cite{troidl2025} is locally retrained with its released architecture and a two-way decode. CB2~\cite{brown2026} uses its predicted masks and the source-described mask-to-split procedure. Because public FlyWire agglomeration affinities are unavailable, its CB2 reconstruction uses native contact area and is treated as a source-faithful adaptation. 3D Part.\ mIoU averages the two native-voxel-weighted side IoUs after optimal side swapping. MIN is the smaller matched-side IoU. S@10 is the fraction with at most 10\% mislabeled native-supervoxel mass and a predicted minority side containing at least 1\% of the total mass. Coverage is the scorer-valid output rate, with NO-OP valid by construction. FA-3D Part.\ mIoU is the primary endpoint and assigns a value of 0 to invalid outputs, reflecting both partition quality and output reliability. The secondary metrics use the common scorer-valid intersection. Contact-valid means $|\mathcal C|>0$. CC-Auto is CPU-only, with median and 95th percentile inference times of 0.30 and 5.65 seconds on MICrONS, and 0.11 and 1.32 seconds on FlyWire.
\begin{table*}[t]
\centering
\caption{3D-native automatic false-merge correction. 
Coverage and FA-3D Part.\ mIoU use the eligible TEST cohort. Eligibility requires a valid native two-way historical reference. The remaining metrics use the intersection of scorer-valid outputs across all applicable displayed methods. $M\!\rightarrow\!FW$ denotes zero-shot transfer. Best/second-best are \textbf{bold}/\underline{underlined}.}
\label{tab:main_results}
\vspace{+0.05in}

\fontsize{7.0}{7.5}\selectfont
\renewcommand{\arraystretch}{0.96}
\setlength{\tabcolsep}{0.9pt}

\begin{tabular*}{\textwidth}{
@{\extracolsep{\fill}}
l
ccc
cc
cccc
cc
@{}
}
\toprule
&
\multicolumn{3}{c}{\textbf{Simple baselines}} &
\multicolumn{2}{c}{\textbf{Graph-based}} &
\multicolumn{4}{c}{\textbf{Learned baselines}} &
\multicolumn{2}{c}{\textbf{Ours}}
\\

\cmidrule(lr){2-4}
\cmidrule(lr){5-6}
\cmidrule(lr){7-10}
\cmidrule(lr){11-12}

\textbf{Metric}
&
NO-OP
&
Random
&
Geometric
&
NC-RW
&
NC-WS
&
PAT$_{M\rightarrow FW}$
&
PAT
&
CB2$_{M\rightarrow FW}$
&
CB2
&
CC-Auto$_{M\rightarrow FW}$
&
\textbf{CC-Auto}
\\
\midrule

\multicolumn{12}{l}{
\textbf{MICrONS}\quad
{\normalfont eligible $N=733$; common-success $N=575$}
}
\\
\addlinespace[1pt]

Coverage (\%)
&
100.0
&
100.0
&
100.0
&
99.2
&
99.2
&
--
&
100.0
&
--
&
78.6
&
--
&
99.2
\\

\textbf{FA-3D Part.\ mIoU} $\uparrow$
&
0.358
&
0.546
&
0.566
&
\underline{0.598}
&
0.588
&
--
&
0.532
&
--
&
0.391
&
--
&
\textbf{0.657}
\\

3D Part.\ mIoU $\uparrow$
&
0.356
&
0.552
&
0.573
&
\underline{0.612}
&
0.599
&
--
&
0.538
&
--
&
0.497
&
--
&
\textbf{0.670}
\\

MIN $\uparrow$
&
0.000
&
0.440
&
\underline{0.467}
&
0.440
&
0.415
&
--
&
0.425
&
--
&
0.303
&
--
&
\textbf{0.558}
\\

ARI $\uparrow$
&
0.000
&
0.265
&
0.298
&
\underline{0.380}
&
0.364
&
--
&
0.242
&
--
&
0.193
&
--
&
\textbf{0.458}
\\

VOI $\downarrow$
&
0.797
&
0.592
&
0.567
&
\underline{0.512}
&
0.530
&
--
&
0.579
&
--
&
0.639
&
--
&
\textbf{0.437}
\\

S@10 $\uparrow$
&
0.000
&
0.172
&
0.209
&
\underline{0.362}
&
0.353
&
--
&
0.153
&
--
&
0.151
&
--
&
\textbf{0.423}
\\

\addlinespace[0.5pt]
\multicolumn{12}{@{}l@{}}{\makebox[0pt][l]{\hdashrule[0.5ex]{\textwidth}{0.4pt}{1.2pt 1.6pt}}}\\
\addlinespace[2.5pt]

\multicolumn{12}{l}{
\textbf{FlyWire}\quad
{\normalfont eligible $N=2929$; common-success $N=1559$}
}
\\
\addlinespace[1pt]

Coverage (\%)
&
100.0
&
100.0
&
100.0
&
99.2
&
99.2
&
100.0
&
100.0
&
68.9
&
65.8
&
99.2
&
99.2
\\

\textbf{FA-3D Part.\ mIoU} $\uparrow$
&
0.3650
&
0.5568
&
0.5775
&
0.6067
&
0.6075
&
0.5342
&
0.5414
&
0.3339
&
0.4164
&
\underline{0.6405}
&
\textbf{0.6414}
\\

3D Part.\ mIoU $\uparrow$
&
0.364
&
0.550
&
0.575
&
0.608
&
0.607
&
0.531
&
0.538
&
0.492
&
0.626
&
\underline{0.650}
&
\textbf{0.651}
\\

MIN $\uparrow$
&
0.000
&
0.427
&
0.462
&
0.424
&
0.421
&
0.409
&
0.394
&
0.342
&
0.485
&
\textbf{0.503}
&
\underline{0.502}
\\

ARI $\uparrow$
&
0.000
&
0.259
&
0.301
&
0.372
&
0.374
&
0.233
&
0.248
&
0.171
&
0.389
&
\underline{0.439}
&
\textbf{0.440}
\\

VOI $\downarrow$
&
0.770
&
0.579
&
0.545
&
0.495
&
0.497
&
0.559
&
0.564
&
0.583
&
0.458
&
\textbf{0.436}
&
\underline{0.437}
\\

S@10 $\uparrow$
&
0.000
&
0.153
&
0.207
&
0.371
&
0.375
&
0.144
&
0.185
&
0.108
&
0.352
&
\underline{0.430}
&
\textbf{0.431}
\\

\bottomrule
\end{tabular*}

\vspace{2pt}

    \vspace{-0.25cm}
\end{table*}
\begin{figure}[t]
    \centering
    \includegraphics[width=\columnwidth]{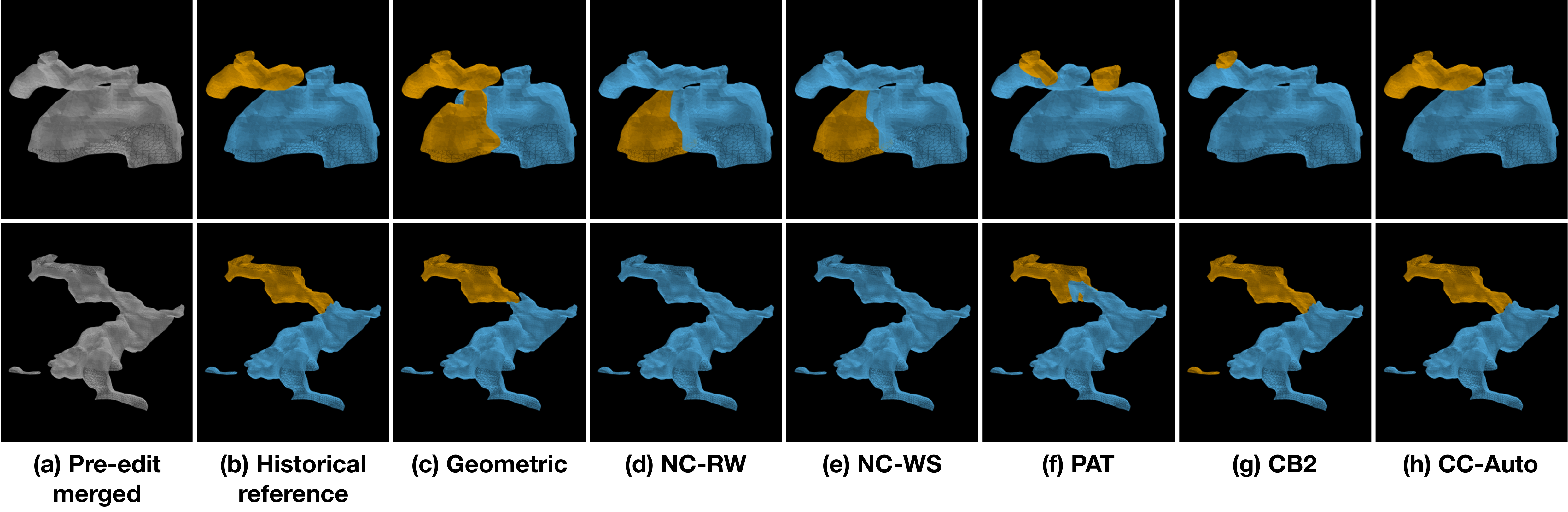}
    \vspace{-0.8cm}
    \caption{Qualitative 3D-native correction results. Predictions are shown against the historical reference.}
    \label{fig:qualitative}
    \vspace{-0.25cm}
\end{figure}

\noindent\textbf{Automatic correction and transfer.}\ 
Table~\ref{tab:main_results} shows that CC-Auto achieves the highest FA-3D Part.\ mIoU on both MICrONS (0.6574) and FlyWire (0.6414) datasets (qualitative examples in Figure~\ref{fig:qualitative}), exceeding the strongest graph-based control by 0.059 (95\% CI [$0.038$, $0.080$]) and 0.034 ([$0.023$, $0.045$]), respectively, while also outperforming the learned baselines. PAT has full scorer-valid coverage, so CC-Auto's advantage over PAT is not driven by invalid-output penalties. The MICrONS-trained CC-Auto model achieves 0.6405 on FlyWire, compared to 0.6414 for in-domain training, indicating that the structural ranker transfers with almost no loss. Excluding only operations unavailable to CB2 because of missing benchmark-side inputs leaves CC-Auto ahead at 0.659 vs.\ 0.398 on MICrONS and 0.642 vs.\ 0.446 on FlyWire.

\noindent\textbf{Localization robustness.}\ 
Displacing the recorded interface cue in five sampled 3D directions changes FA-3D Part.\ mIoU by at most 0.009 for perturbations up to 250\,nm on either dataset. Larger degradation appears at 500\,nm as the original contact leaves the fixed candidate neighborhood.
\begin{table}[t]
\centering
\caption{Contact-selection ablation and bounded search. 
3D Part.\ mIoU on contact-valid TEST operations; $^\dagger$ denotes reference-best reachable or best-in-panel quality.}
\label{tab:ablation_search}
\vspace{+0.05in}

\fontsize{7.0}{7.4}\selectfont
\renewcommand{\arraystretch}{0.94}
\setlength{\tabcolsep}{1.3pt}
\begin{tabular*}{\columnwidth}{
@{\extracolsep{\fill}}
lccc
@{\hspace{8pt}}
lcc
@{}
}
\toprule
\multicolumn{4}{c}{\textbf{(a) Selector $\times$ solver}}
&
\multicolumn{3}{c}{\textbf{(b) Bounded search}}
\\
\cmidrule(lr){1-4}
\cmidrule(lr){5-7}
\textbf{Selector}
& \textbf{GC}
& \textbf{RW}
& \textbf{WS}
& \textbf{Condition}
& \textbf{MICrONS}
& \textbf{FlyWire}
\\
\midrule
\multicolumn{4}{l}{\textit{MICrONS} ($N=727$)}
& H\_FIXED base & 0.641 & 0.639
\\
Nearest & 0.557 & 0.603 & 0.593
& Best-single$^\dagger$ & 0.843 & 0.798
\\
Auto & \textbf{0.663} & \textbf{0.654} & \textbf{0.643}
& B45-pair$^\dagger$ & \textbf{0.866} & \textbf{0.837}
\\
\multicolumn{4}{l}{\textit{FlyWire} ($N=2906$)}
& Best-in-$K5^\dagger$ & 0.826 & 0.800
\\
Nearest & 0.562 & 0.612 & 0.612
& & &
\\
Auto & \textbf{0.646} & \textbf{0.637} & \textbf{0.626}
& & &
\\
\bottomrule
\end{tabular*}
\vspace{2pt}

\end{table}

\begin{figure}[t]
    \centering
    \includegraphics[width=\columnwidth]{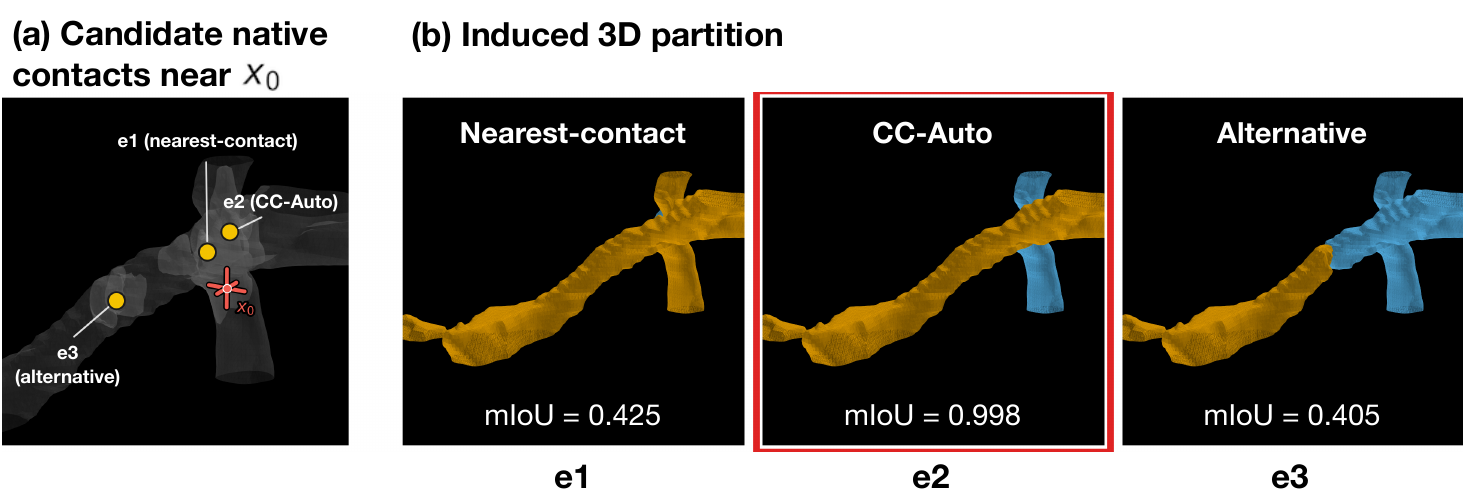}
    \vspace{-0.9cm}
    \caption{Effect of contact selection.
    The red marker in (a) is the localized cue $x_0$. This example shows how nearby contacts induce different partitions under the same graph-cut setup. $e_2$ is selected by CC-Auto.}
    \label{fig:intervention}
    \vspace{-0.25cm}
\end{figure}

\noindent\textbf{Contact selection and learned weighting.}\ 
Table~\ref{tab:ablation_search}(a) isolates contact selection from the downstream solver. Figure~\ref{fig:intervention} illustrates the resulting partition differences. Auto terminals are selected using graph-cut-induced partitions and reused unchanged with RW/WS. Auto improves mean 3D Part. mIoU over nearest-contact selection under all three solvers on both datasets. All six paired component-bootstrap 95\% confidence intervals exclude zero.

On the full eligible TEST cohorts (MICrONS/FlyWire), smaller-side-only, cut-size-only, and equal-weight scoring achieve FA-3D Part.\ mIoU of 0.6359/0.6325, 0.4402/0.4639, and 0.6478/0.6272, respectively, compared with 0.6574/\allowbreak0.6414 for learned weighting. H\_FIXED achieves 0.6357/\allowbreak0.6334. To isolate learning, equal-weight scoring uses $\theta=(1,1)$ with the same features $\phi$, connectivity tier, and solver as CC-Auto. Learned-minus-equal gains are $0.0096$ (95\% CI [$-0.0005$, $0.0189$]) on MICrONS and $0.0142$ ([$0.0093$, $0.0191$]) on FlyWire. Learning improves mean performance, although the MICrONS interval includes zero.

\noindent\textbf{Bounded search.}\ 
Table~\ref{tab:ablation_search}(b) shows that paired-contact search increases reference-best reachable mIoU from 0.843/0.798 for single contacts to 0.866/0.837 for the H\_FIXED-plus-B45 pool on MICrONS/FlyWire. The base-plus-five panel retains 0.826/0.800. On 725 MICrONS and 2887 FlyWire TEST panels, PairLogit improves NEAR05@2 over fixed structural ordering from 0.539 to 0.621 and from 0.605 to 0.680, respectively, with zero-shot transfer to FlyWire. Panel membership is unchanged. These results measure the quality and prioritization of what is reachable for proofreading or downstream selection, rather than the accuracy of automatic correction.

\vspace{-0.25cm}
\section{Conclusions}
\vspace{-0.15cm}
\label{sec:conclusions}
ContactCut targets localized false-merge correction through 3D-native partition-aware contact selection. Partition-aware structural selection significantly improves performance over nearest-contact placement. A learned ranker offers further gains and transfers effectively from MICrONS to FlyWire. Bounded paired-contact search expands reachable correction quality, while a learned panel ranker efficiently prioritizes near-oracle alternatives. These results underscore the importance of contact selection in translating localized error cues into accurate, executable 3D-native false-merge corrections.

\vspace{-0.25cm}
\section{Compliance with ethical standards}
\vspace{-0.15cm}
\label{sec:ethics}

This retrospective computational study used publicly available connectomics data and involved no new animal experiments. MICrONS animal procedures were approved by the IACUC at Baylor College of Medicine and the Allen Institute for Brain Science. FlyWire uses the published FAFB \textit{Drosophila melanogaster} dataset.

\vspace{-0.25cm}
\section{Acknowledgments}
\vspace{-0.15cm}
\label{sec:acknowledgments}

This research was supported in part by the U.S. National Science Foundation through grant OAC-2104158, and the U.S. National Institutes of Health through grant 7R01HL177814. The authors have no relevant financial or non-financial interests to disclose.




\vspace{-0.25cm}
\bibliographystyle{IEEEbib}
\bibliography{refs-abbv}

\end{document}